\documentclass[10pt,twocolumn,letterpaper]{article}

\usepackage{wacv}
\usepackage{times}
\usepackage{epsfig}

\usepackage[T1]{fontenc}
\usepackage{microtype}
\usepackage{graphicx}
\usepackage{subfigure}
\usepackage{capt-of}
\usepackage{color}
\usepackage{cite}
\usepackage{tikz}
\usepackage[accsupp]{axessibility}  
\usepackage{amsmath,amssymb,amsfonts,dsfont,pifont,bm,bbm,mathrsfs,mathtools}
\usepackage{algorithm,algpseudocode,listings}
\usepackage{booktabs,multirow,adjustbox,diagbox}
\usepackage{threeparttable}
\usepackage{xspace}
\ifwacvfinal
\usepackage[breaklinks=true,bookmarks=false]{hyperref}
\else
\usepackage[pagebackref=true,breaklinks=true,colorlinks,bookmarks=false]{hyperref}
\fi
\usepackage{cleveref}  
\crefname{section}{Sec.}{Secs.}
\Crefname{section}{Section}{Sections}
\crefname{table}{Tab.}{Tabs.}
\Crefname{table}{Table}{Tables}
\crefname{figure}{Fig.}{Figs.}
\Crefname{figure}{Figure}{Figures}
\crefname{equation}{Eq.}{Eqs.}
\Crefname{equation}{Equation}{Equations}
\def\wacvPaperID{583} 

\wacvalgorithmstrack   

\wacvfinalcopy 

\newcommand{\tocite}[1]{\textcolor{red}{[TO CITE]}}

\newcommand{\method}{SAFECount\xspace}
\newcommand{\supp}{\textit{Supplementary Material}\xspace}
\newcommand{\tabincell}[2]{\begin{tabular}{@{}#1@{}}#2\end{tabular}}

\newcommand{\fs}{\bm{f_S}}
\newcommand{\fq}{\bm{f_Q}}
\newcommand{\Rraw}{\bm{R_0}}
\newcommand{\Rnorm}{\bm{R}}
\newcommand{\fweight}{\bm{f_R}}
\newcommand{\fout}{\bm{f_Q'}}

\begin{document}

\title{Few-shot Object Counting with Similarity-Aware Feature Enhancement}

\author{Zhiyuan You$^1$,\quad Kai Yang$^2$,\quad Wenhan Luo$^3$,\quad Xin Lu$^2$,\quad Lei Cui$^4$,\quad Xinyi Le$^{1*}$
\\
$^1$Shanghai Jiao Tong University,\quad $^2$SenseTime,\quad $^3$Tencent,\quad $^4$Tsinghua University
\\
{\tt\small zhiyuanyou@foxmail.com, \{yangkai, luxin\}@sensetime.com, whluo.china@gmail.com}
\\
{\tt\small cuil19@mails.tsinghua.edu.cn, lexinyi@sjtu.edu.cn, $^*$Corresponding Author}
}

\twocolumn[{
	\renewcommand\twocolumn[1][]{#1}
	\maketitle
	\thispagestyle{empty}
	\vspace{-18pt}
	\begin{center}
		\includegraphics[width=0.95\linewidth]{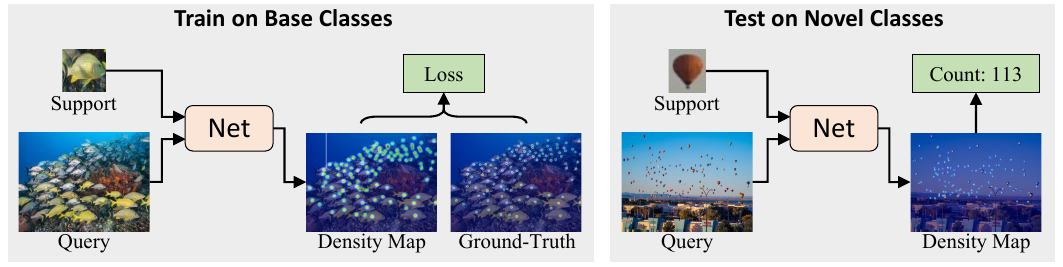}
		\vspace{-3pt}
		\captionof{figure}{
    		\textbf{Illustration of few-shot object counting}, where we would like to find how many exemplar objects described by \textit{a few} support images occur in the query image.
            Besides the objects included in the training phase, we also expect the model to handle novel classes at the test stage \textit{without retraining}.
		}
		\label{fig:setting}
		\vspace{5pt}
	\end{center}
}]

\begin{abstract}

This work studies the problem of few-shot object counting, which counts the number of exemplar objects (\textit{i.e.}, described by one or several support images) occurring in the query image.
The major challenge lies in that the target objects can be densely packed in the query image, making it hard to recognize every single one.
To tackle the obstacle, we propose a novel learning block, equipped with a similarity comparison module and a feature enhancement module.
Concretely, given a support image and a query image, we first derive a score map by comparing their projected features at every spatial position.
The score maps regarding all support images are collected together and normalized across both the exemplar dimension and the spatial dimensions, producing a reliable similarity map.
We then enhance the query feature with the support features by employing the developed point-wise similarities as the weighting coefficients.
Such a design encourages the model to inspect the query image by \textit{focusing more on the regions akin to the support images, leading to much clearer boundaries between different objects}.
Extensive experiments on various benchmarks and training setups suggest that we surpass the state-of-the-art methods by a sufficiently large margin.
For instance, on a recent large-scale FSC-147 dataset, we surpass the state-of-the-art method by improving the mean absolute error from 22.08 to 14.32 (35\%$\uparrow$). 
Code has been released \href{https://github.com/zhiyuanyou/SAFECount}{here}. 

%
%

\end{abstract}
\vspace{-12pt}

\section{Introduction}\label{sec:introduction}


Object counting~\cite{ucsd, mall}, which aims at investigating how many times a certain object occurs in the query image, has received growing attention due to its practical usage~\cite{pdem, mcnn, blobs, liu2021cross}.
Most existing studies assume that the object to count at the test stage is covered by the training data~\cite{liu2021cross, song2021rethinking, zhang2021cross, mcnn, penguin, lpn, hlcnn}. 
As a result, each learned model can only handle a specific object class, greatly limiting its application.
%

To alleviate the generalization problem, few-shot object counting (FSC) is recently introduced~\cite{famnet}.
Instead of pre-defining a common object that is shared by all training images, FSC allows users to customize the object of their own interests with a few support images, as shown in \cref{fig:setting}. 
In this way, we can use a single model to unify the counting of various objects, and even adapt the model to novel classes (\textit{i.e.}, unseen in the training phase) without any retraining.

\begin{figure*}[t]
    \centering
    \includegraphics[width=0.95\linewidth]{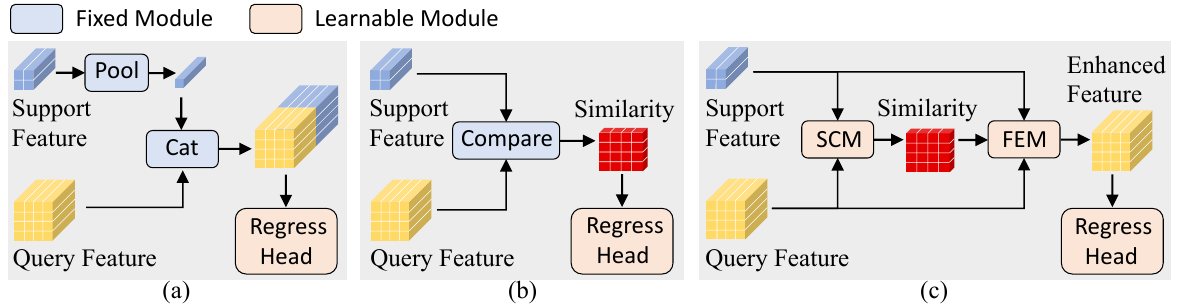}
    \vspace{-3pt}
    \caption{
        \textbf{Concept comparison} between our method and existing alternatives.
        (a) Feature-based approach~\cite{gmn}, where the query feature is concatenated with the pooled support feature for regression.
        (b) Similarity-based approach~\cite{cfocnet, famnet}, where a similarity map is developed from raw features for regression.
        (c) Our proposed \textit{similarity-aware feature enhancement} block, consisting of a similarity comparison module (SCM) and a feature enhancement module (FEM).
        Concretely, the reliable feature similarity developed by SCM is exploited as the guidance of FEM to enhance the query feature with the support feature.
        %
        %
        The details of SCM and FEM can be found in \cref{subsec:core-block} and \cref{fig:framework}. 
    }
    \label{fig:compare}
    \vspace{-15pt}
\end{figure*}

A popular solution to FSC is to first represent both the exemplar object (\textit{i.e.} the support image) and the query image with expressive features, and then pinpoint the candidates via analyzing the feature correlation~\cite{gmn, cfocnet, famnet}.
Active attempts roughly fall into two folds.
One is feature-based~\cite{gmn}, as shown in \cref{fig:compare}a, where the pooled support feature is concatenated onto the query feature, followed by a regress head to recognize whether the two features are close enough.
However, the spatial information of the support image is omitted by pooling, leaving the feature comparison unreliable.
The other is similarity-based~\cite{cfocnet, famnet}, as shown in \cref{fig:compare}b, where a similarity map is developed from raw features as the regression object.
Nevertheless, the similarity is far less informative than feature, making it hard to identify clear boundaries between objects (see \cref{fig:relation}).
Accordingly, the counting performance heavily deteriorates when the target objects are densely packed in the query image, like the shoal of fish in \cref{fig:setting}.

In this work, we propose a Similarity-Aware Feature Enhancement block for object Counting (\method).
As discussed above, \textit{feature is more informative while similarity better captures the support-query relationship}. 
Our novel block adequately integrates both of the advantages by exploiting similarity as a guidance to enhance the features for regression.
Intuitively, the enhanced feature not only carries the rich semantics extracted from the image, but also gets aware of which regions within the query image are similar to the exemplar object.
Specifically, we come up with a similarity comparison module (SCM) and a feature enhancement module (FEM), as illustrated in \cref{fig:compare}c.
On one hand, different from the naive feature comparison in \cref{fig:compare}b, our SCM learns a feature projection, then performs a comparison on the projected features to derive a score map. 
This design helps select from features the information that is most appropriate for object counting.
After the comparison, we derive a reliable similarity map by collecting the score maps with respect to all support images (\textit{i.e.}, few-shot) and normalizing them along both the exemplar dimension and the spatial dimensions.
On the other hand, the FEM takes the point-wise similarities as the weighting coefficients, and fuses the support features into the query feature.
Such a fusion is able to make the enhanced query feature focus more on the regions akin to the exemplar object defined by support images, facilitating more precise counting.

Experimental results on a very recent large-scale FSC dataset, FSC-147~\cite{famnet}, and a car counting dataset, CARPK~\cite{lpn}, demonstrate our \textit{substantial improvement} over state-of-the-art methods.
Through visualizing the intermediate similarity map and the final predicted density map, we find that our \method substantially benefits from the clear boundaries learned between objects, even when they are densely packed in the query image.

\section{Related Work}\label{sec:related}

\vspace{2pt}\noindent\textbf{Class-specific object counting}
counts objects of a specific class, such as people~\cite{liu2021cross, song2021rethinking, zhang2021cross, mcnn}, animals~\cite{penguin}, cars~\cite{lpn}, among which crowd counting has been widely explored.
For this purpose, traditional methods~\cite{leibe2005pedestrian, stewart2016end, wang2011automatic} count the number of people occurring in an image through person detection.
However, object detection is not particularly designed for the counting task and hence shows unsatisfying performance when the crowd is thick.
To address this issue, recent work~\cite{wan2019residual} employs a deep model to predict the density map from the crowd image, where the sum over the density map gives the counting result~\cite{lempitsky2010learning}.
Based on this thought, many attempts have been made to handle more complicated cases~\cite{yan2019perspective, rpnet, switch_cnn, mcnn, ranjan2018iterative, sam2018top, shi2018crowd, sindagi2017generating, xiong2017spatiotemporal, crowd_cnn, liu2018decidenet}.
%
%
%
%
Some recent studies~\cite{song2021rethinking, glf} propose effective loss functions that help predict the position of each person precisely.
%
%
However, all of these methods can only count objects regarding a particular class (\textit{e.g.}, person), making them hard to generalize.
There are also some approaches targeting counting objects of multiple classes~\cite{iep_counting, blobs, michel2022class, xu2021dilated}.
In particular, Stahl~\textit{et al.}~\cite{iep_counting} propose to divide the query image into regions and regress the counting results with the inclusion-exclusion principle.
Laradji~\textit{et al.}~\cite{blobs} formulate counting as a segmentation problem for better localization.
Michel~\textit{et al.}~\cite{michel2022class} detect target objects and regress multi-class density maps simultaneously.
Xu~\textit{et al.}~\cite{xu2021dilated} mitigate the mutual interference across various classes by proposing category-attention module.
Nevertheless, they still can not handle the object classes beyond the training data.

\vspace{2pt}\noindent\textbf{Few-shot object counting (FSC)}
has recently been proposed~\cite{gmn, cfocnet, famnet} and presents a much stronger generalization ability.
Instead of pre-knowing the type of object to count, FSC allows users to describe the exemplar object of their own interests with \textit{one or several} support images.
This setting makes the model highly flexible in that it does not require the test object to be covered by the training samples.
In other words, a well-learned model could easily make inferences on novel classes (\textit{i.e.}, unseen in the training phase) as long as the support images are provided.
To help the model dynamically get adapted to an arbitrary class, a great choice is to compare the object and the query image in feature space~\cite{gmn, cfocnet, famnet}.
GMN~\cite{gmn} pools the support feature, and concatenates the pooling result onto the query feature, then learns a regression head for point-wise feature comparison.
However, the comparison built on concatenation is not as reliable as the similarity~\cite{cfocnet}.
Instead, CFOCNet~\cite{cfocnet} first performs feature comparison with dot production, and then regresses the density map from the similarity map derived before.
FamNet~\cite{famnet} further improves the reliability of the similarity map through multi-scale augmentation and test-time adaptation.
But similarities are far less informative than features, hence regressing from the similarity map fails to identify clear boundaries between the densely packed objects.
In this work, we propose a similarity-aware feature enhancement block, which integrates the advantages of both features and similarities.


\vspace{2pt}\noindent\textbf{Few-shot learning} has received popular attention in the past few years thanks to its high data efficiency~\cite{maml, wertheimer2021few, fan2020few, wu2020multi, yang2021mining, zhang2019canet}.
The rationale behind this is to adapt a well-trained model to novel test data (\textit{i.e.}, having a domain gap to the training data) with a few support samples.
In the field of image classification~\cite{maml, wertheimer2021few}, MAML~\cite{maml} proposes to fit parameters to novel classes at the test stage using a few steps of gradient descent.
FRN~\cite{wertheimer2021few} formulates few-shot classification as a reconstruction problem.
As for object detection~\cite{fan2020few, wu2020multi}, Fan~\textit{et al.}~\cite{fan2020few} exploit the similarity between the input image and the support images to detect novel objects. 
Wu~\textit{et al.}~\cite{wu2020multi} create multi-scale positive samples as the object pyramid for prediction refinement.
When the case comes to semantic segmentation~\cite{yang2021mining, zhang2019canet}, CANet~\cite{zhang2019canet} iteratively refines the segmentation results by comparing the query feature and the support feature. 
Yang~\textit{et al.}~\cite{yang2021mining} aim to alleviate the problem of feature undermining and enhance the embedding of novel classes.
In this work, we explore the usage of few-shot learning on the object counting task.

\section{Method}\label{sec:method}

\begin{figure*}[t]
    \centering
    \includegraphics[width=0.95\linewidth]{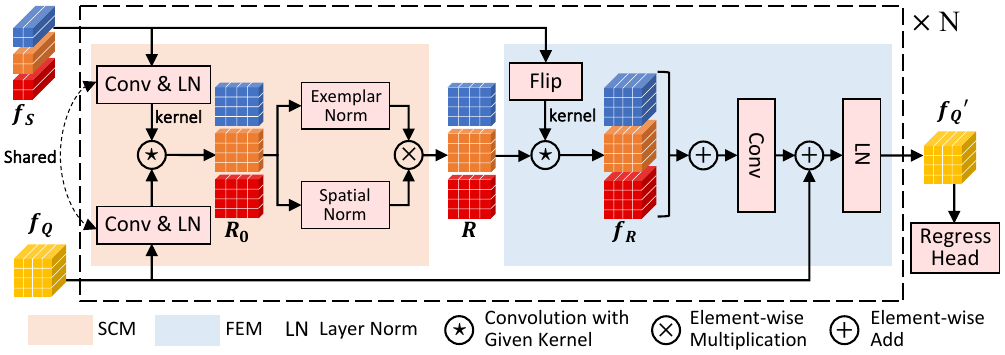}
    \vspace{-3pt}
    \caption{
        \textbf{Illustration of the similarity-aware feature enhancement block} under the 3-shot case.
        Given features, $\fs$, $\fq$, that are extracted from the support images and the query image respectively, the similarity comparison module (SCM) first develops a score map, $\Rraw$, by comparing the projected features, and then produces a similarity map, $\Rnorm$, via normalizing $\Rraw$ along both the exemplar dimension and the spatial dimensions.
        Here, feature projection is implemented with a $1\times1$ convolution.
        The following feature enhancement module (FEM) weights $\fs$ with $\Rnorm$ to derive a similarity-weighted feature, $\fweight$, and manages to fuse $\fweight$ into $\fq$ as a feature enhancement. 
        Such a block can be stacked for multiple times in the training framework. 
    }
    \label{fig:framework}
    \vspace{-15pt}
\end{figure*}

\subsection{Preliminaries}\label{subsec:preliminaries}

Few-shot object counting (FSC)~\cite{famnet} aims to count the number of exemplar objects occurring in a query image with only a few support images describing the exemplar object. 
In FSC, object classes are divided into base classes $\mathcal{C}_b$ and novel classes $\mathcal{C}_n$, where $\mathcal{C}_b$ and $\mathcal{C}_n$ have no intersection. 
For each query image from $\mathcal{C}_b$, both a few support images and the ground-truth density map are provided. 
While, for query images from $\mathcal{C}_n$, only a few support images are available. 
FSC aims to count exemplar objects from $\mathcal{C}_n$ using only a few support images by leveraging the generalization knowledge from $\mathcal{C}_b$. 
If we denote the number of support images for one query image as $K$, the task is called $K$-shot FSC.

\subsection{Similarity-Aware Feature Enhancement}\label{subsec:core-block}

\vspace{2pt}\noindent\textbf{Overview.}
\cref{fig:framework} illustrates the core block in our framework, termed as the \textit{similarity-aware feature enhancement} block.
We respectively denote the support feature and the query feature as $\fs \in \mathbb{R}^{K \times C \times H_S \times W_S}$ and $\fq \in \mathbb{R}^{C \times H_Q \times W_Q}$, where $K$ is the number of support images. 
The similarity comparison module (SCM) first projects $\fs$ and $\fq$ to a comparison space, then compares these projected features at every spatial position, deriving a score map, $\Rraw$. 
Then, $\Rraw$ is normalized along both the exemplar dimension and the spatial dimensions, resulting in a reliable similarity map, $\Rnorm$. 
The following feature enhancement module (FEM) first obtains the similarity-weighted feature, $\fweight$, by weighting $\fs$ with $\Rnorm$, and then manages to fuse $\fweight$ into $\fq$, producing the enhanced feature, $\fout$.
By doing so, the features regarding the regions similar to the support images are ``highlighted'', which could help the model get distinguishable borders between densely packed objects. 
Finally, the density map is regressed from $\fout$.

\vspace{2pt}\noindent\textbf{Similarity Comparison Module (SCM).}
As discussed above, similarity can better characterize how a particular image region is alike the exemplar object.
However, we find that the conventional feature comparison approach (\textit{i.e.}, using the vanilla dot production) used in prior arts~\cite{cfocnet, famnet} is not adapted to fit the FSC task.
By contrast, our proposed SCM develops a reliable similarity map from the input features with the following three steps.

\textit{Step-1: Learnable Feature Projection.}
Before performing feature comparison, $\fs$ and $\fq$ are first projected to a comparison space via a $1\times1$ convolutional layer. 
This projection asks the model to automatically select suitable information from the features.
We also add a shared layer normalization after the projection to make these two features subject to the same distribution as much as possible.
%

\textit{Step-2: Feature Comparison.}
The point-wise feature comparison is realized with convolution.
In particular, we convolve the projected $\fq$ with the projected $\fs$ as kernels, which gives us the score map, $\Rraw \in \mathbb{R}^{K \times 1 \times H_Q \times W_Q}$, as
\begin{align}
  \scriptsize
  \Rraw = \mathtt{conv} ( g(\bm{\fq}), \text{kernel} = g(\bm{\fs})), \label{equ:relation}
\end{align}
where $g(\cdot)$ denotes the feature projection described in \textit{Step-1}, \textit{i.e.}, a $1\times1$ convolutional layer followed layer normalization.
%

\textit{Step-3: Score Normalization.}
The values of the score map, $\Rraw$, are normalized to a proper range to avoid some unusual (\textit{e.g.}, too large) entries from dominating the learning.
Here, we propose Exemplar Normalization (ENorm) and Spatial Normalization (SNorm). 
On the one hand, ENorm normalizes $\Rraw$ along the exemplar dimension as
\begin{align}
  \bm{R_{EN}} = \mathtt{softmax}_{dim=0}(\frac{\Rraw}{\sqrt{H_S W_S C}}), 
\end{align}
where $\mathtt{softmax}_{dim}(\cdot)$ is the softmax layer operated along a specific dimension. 
%
%
%
%
On the other hand, $\Rraw$ is also normalized along the spatial dimensions (\textit{i.e.}, the height and width) with SNorm, as
\begin{align}
  \bm{R_{SN}}  = \frac{\exp(\Rraw / \sqrt{H_S W_S C})}{\max_{dim=(2,3)}(\exp(\Rraw / \sqrt{H_S W_S C}))}, 
\end{align}
where $\max_{dim}(\cdot)$ finds the maximum value from the given dimensions. 
After SNorm, the score value of the most support-relevant position would be 1, and others would be among $[0,1]$.
Finally, the similarity map, $\Rnorm$, is obtained from $\bm{R_{EN}}$ and $\bm{R_{SN}}$ with
\begin{align}
    \Rnorm = \bm{R_{EN}} \otimes \bm{R_{SN}} \in \mathbb{R}^{K \times 1 \times H_Q \times W_Q},
\end{align}
where $\otimes$ is the element-wise multiplication.
The studies of the effect of ENorm and SNorm can be found in \cref{subsec:ablation}.


\vspace{2pt}\noindent\textbf{Feature Enhancement Module (FEM).}
Recall that, compared to similarity, feature is more informative in representing the image yet less accurate in capturing the support-query relationship.
To take sufficient advantages of both, we propose to use the similarity developed by SCM as the guidance for feature enhancement.
Specifically, our FEM integrates the support feature, $\fs$, into the query feature, $\fq$, with similarity values in $\Rnorm$ as the weighting coefficients.
In this way, the model can inspect the query image by paying more attention to the regions that are akin to the support images.
This module consists of the following two steps.

\textit{Step-1: Weighted Feature Aggregation.}
In this step, we aggregate the support feature, $\fs$, by taking the point-wise similarity, $\Rnorm$, into account. 
Namely, the feature point corresponding to a higher similarity score should have larger voice to the final enhancement.
Such a weighted aggregation is implemented with convolution, which outputs the similarity-weighted feature,
\begin{align}
  \fweight' & = \mathtt{conv}(\bm{R}, {\rm kernel} = \mathtt{flip}(\bm{f_S})) \in \mathbb{R}^{K \times C \times H_Q \times W_Q}, \\
  \fweight & = \mathtt{sum}_{dim=0}(\fweight') \in \mathbb{R}^{C \times H_Q \times W_Q},  \label{equ:f_R}
\end{align}
where $\mathtt{sum}_{dim}(\cdot)$ accumulates the input tensor along specific dimensions, 
$\mathtt{flip}(\cdot)$ denotes the flipping operation, which flips the input tensor both horizontally and vertically.
%
%
Flipping helps $\fweight'$ preserve the spatial structure of $\fs$.
The intuitive illustration and the performance improvement of flipping can be found in \supp.

\textit{Step-2: Learnable Feature Fusion.}
The similarity-weighted feature, $\fweight$, is fused into the query feature, $\fq$, via an efficient network.
It contains a convolutional block and a layer normalization, as shown in \cref{fig:framework}.
Finally, we obtain the enhanced feature, $\fq'$, with
\begin{align}
  \fq' = \mathtt{layer\_norm}(\fq + h(\fweight)) \in \mathbb{R}^{C \times H_Q \times W_Q},  \label{equ:fuse}
\end{align}
where $h(\cdot)$ is implemented with two convolutional layers. 
%

\vspace{2pt} \noindent \textbf{Comparison with Attention}. A classical attention module~\cite{attention_need} involved with \textit{query}, \textit{key}, and \textit{value} (denoted as $q,k,v$) is represented as $\mathtt{similarity}(q, k) v$. 
The key idea is employing the similarity values between $q$ and $k$ as weighting coefficients to aggregate $v$ as an information aggregation. Our \method is similar with the similarity-guided aggregation of existing attention modules. 
However, existing attention modules omit the spatial information as they need to flatten a feature map ($C \times H \times W$) to a collection of feature vectors ($C \times H W$). Instead, in all processes of our SCM and FEM, the feature maps are designed to \textit{maintain their spatial structure} ($C \times H \times W$), which plays a vital role in \textit{learning clear boundaries between objects}. 
The ablation study in \cref{subsec:ablation} confirms our significant advantage over the classical attention module.

\subsection{Training Framework}\label{subsec:framework}

\cref{subsec:core-block} describes the core block of our approach, \method.
In practice, such a block should work together with a feature extractor, which feeds input features into the block, and a regression head, which receives the enhanced feature for object counting.
Moreover, it is worth mentioning that our \method allows stacking itself for continuous performance improvement.
In this part, we will introduce these assistant modules, whose detailed structures are included in \supp.

\vspace{2pt}\noindent\textbf{Feature Extractor.}
When introducing our \method block, we start with the support feature, $\fs$, and the query feature, $\fq$, which are assumed to be well prepared.
Specifically, we use a \textit{fixed} ResNet-18~\cite{resnet} pre-trained on ImageNet~\cite{imagenet} as the feature extractor.
In particular, given a query image, we resize the outputs of the first three stages of ResNet-18 to the same size, $H_Q \times W_Q$, and concatenate them along the channel dimension as the query feature.
Besides, given a support image, which is usually cropped from a large image so as to contain the exemplar object only, the support feature is obtained by applying ROI pooling~\cite{faster_rcnn} on the feature extracted from its parent before cropping.
Here, the ROI pooling size is the size of $\fs$, \textit{i.e.}, $H_S \times W_S$.

\vspace{2pt}\noindent\textbf{Regression Head.}
After getting the enhanced feature, $\bm{f_Q'}$, we convert it to a density map, $\bm{D} \in \mathbb{R}^{H \times W}$, with a regression head.
Following existing methods~\cite{gmn, cfocnet, famnet}, the regression head is implemented with a sequence of convolutional layers, followed by Leaky ReLU activation and bi-linear upsampling.

\vspace{2pt}\noindent\textbf{Multi-block Architecture.}
Recall that our proposed \method block enhances the input query feature, $\fq$, with the support features, $\fs$.
The enhanced feature, $\fq'$, is with exactly the same shape as $\fq$.
As a result, it can be iteratively enhanced simply by stacking more blocks.
The ablation study on the number of blocks can be found in \cref{subsec:ablation}, where we verify that \textit{adding one block is already enough to boost the performance substantially}.

\vspace{2pt}\noindent\textbf{Objective Function.}
Most counting datasets are annotated with the coordinates of the target objects within the query image~\cite{ucsd, mall, mcnn}.
However, directly regressing the coordinates is hard~\cite{lempitsky2010learning, mcnn}.
Following prior work~\cite{famnet}, we generate the ground-truth density map, $\bm{D_{GT}} \in \mathbb{R}^{H \times W}$, from the labeled coordinates, using Gaussian smoothing with adaptive window size.
Our model is trained with the MSE loss as
\begin{align}
    \mathcal{L} = \frac{1}{H \times W} ||\bm{D} - \bm{D_{GT}}||_2^2.  \label{equ:mse_loss}
\end{align}

\section{Experiments} \label{sec:exp}

\subsection{Metrics and Datasets}

\vspace{2pt}\noindent \textbf{Metrics.}  We choose Mean Absolute Error (MAE) and Root Mean Squared Error (RMSE) to measure the performance of counting methods following~\cite{pdem, famnet}:
\begin{equation}
\begin{aligned}
    MAE = & \frac{1}{N_Q} \sum_{i=1}^{N_Q} |C^i - C^i_{GT}|, \\ RMSE = & \sqrt{\frac{1}{N_Q} \sum_{i=1}^{N_Q} (C^i - C^i_{GT})^2},
\end{aligned}
\label{eq:metric}
\end{equation}
where $N_Q$ is the number of query images, $C^i$ and $C^i_{GT}$ are the predicted and ground-truth count of the $i^{th}$ query image, respectively.

\vspace{2pt}\noindent \textbf{FSC-147}.  FSC-147~\cite{famnet} is a multi-class, 3-shot FSC dataset with $147$ classes and $6135$ images. Each image has $3$ support images to describe the target objects. Note that the training classes share no intersection with the validation classes and test classes. The training set contains $89$ classes, while validation set and test set both contain another disjoint $29$ classes. The number of objects per image varies extremely from $7$ to $3701$ with an average of $56$.

\vspace{2pt}\noindent \textbf{Cross-validation of FSC-147.}  
In original FSC-147~\cite{famnet}, the dataset split and the shot number are both fixed, while other few-shot tasks including classification~\cite{wertheimer2021few}, detection~\cite{fan2020few}, and segmentation~\cite{yang2021mining} all contain multiple dataset splits and shot numbers. 
Therefore, we propose to evaluate FSC methods with multiple dataset splits and shot numbers by incorporating FSC-147 and cross-validation. 
Specifically, we split all images in FSC-147 to 4 folds, whose class indices, class number, and image number are shown in \cref{tab:fold}. 
The class indices ranging from 0 to 146 are obtained by sorting the class names of all 147 classes.
Note that these 4 folds share no common classes. 
When fold-$i$ $(i=0,1,2,3)$ serves as the test set, the remaining 3 folds form the training set. 
Also, we evaluate FSC methods in both 1-shot and 3-shot cases. 
For 3-shot case, the original three support images in FSC-147 are used. 
For 1-shot case, we randomly sample one from the original three support images.

\begin{table}[t]
\setlength\tabcolsep{5pt}
\centering
\scriptsize
\caption{Statistics of the four fold splits from FSC-147~\cite{famnet}.}
\begin{tabular}{cccc}
\toprule
Fold & Class Indices & \#Classes & \#Images  \\
\midrule
0 & 0-35         & 36 & 2033 \\
1 & 36-72        & 37 & 1761 \\
2 & 73-109       & 37 & 1239 \\
3 & 110-146      & 37 & 1113 \\
\bottomrule
\end{tabular}
\label{tab:fold}
\vspace{-10pt}
\end{table}

\vspace{2pt}\noindent \textbf{CARPK.} A car counting dataset, CARPK~\cite{lpn}, is used to test our model’s ability of cross-dataset generality. CARPK contains $1448$ images and nearly $90,000$ cars from a drone perspective. These images are collected in various scenes of $4$ different parking lots. The training set contains $3$ scenes, while another scene is used for test.

\subsection{Class-agnostic Few-shot Object Counting} \label{sec:fsc_147}
Our method is evaluated on the FSC dataset FSC-147~\cite{famnet} under the original setting and the cross-validation setting. 

\vspace{2pt}\noindent \textbf{Setup}. The sizes of the query image, the query feature map, and the support feature map, $H \times W$, $H_Q \times W_Q$, and $H_S \times W_S$, are selected as $512 \times 512$, $128 \times 128$, and $3 \times 3$, respectively. The dimension of the projected features are set as 256. The multi-block number is set as $4$. The model is trained with Adam optimizer~\cite{adam} for $200$ epochs with batch size $8$. The learning rate is set as 2e-5 initially, and it is dropped by $0.25$ every $80$ epochs. 

\begin{figure*}[!ht]
    \centering
    \includegraphics[width=0.9\linewidth]{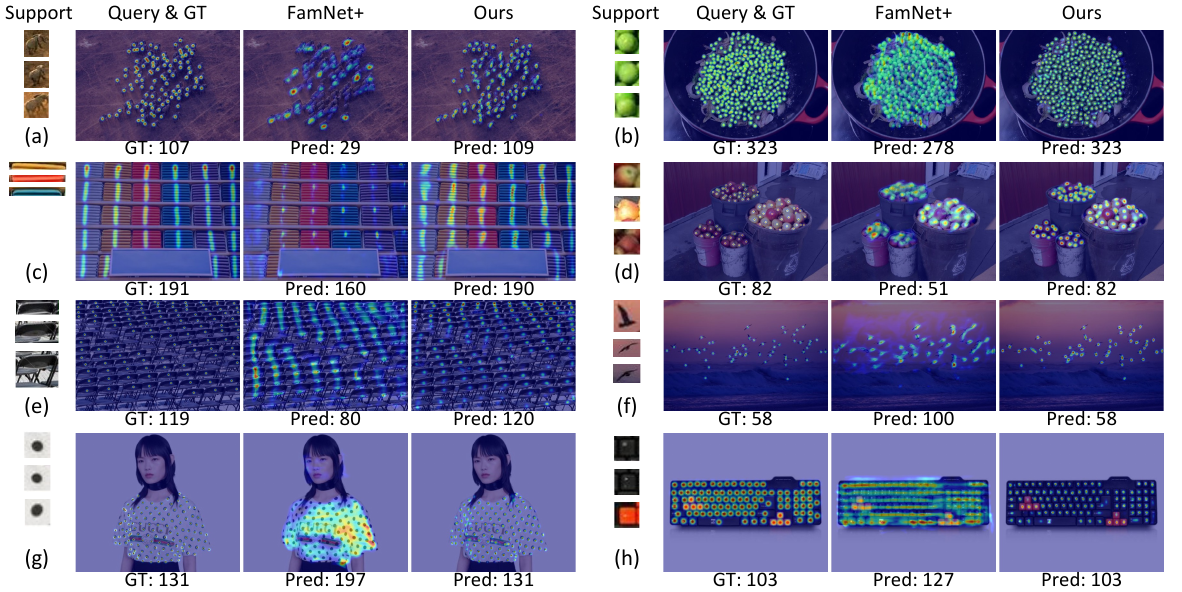}
    \vspace{-2pt}
    \captionof{figure}{
        \textbf{Qualitative results} on the FSC-147 dataset~\cite{famnet} under the 3-shot case.
        From left to right: support images, query image overlaid by the ground-truth density map, predicted density map by FamNet+~\cite{famnet}, and our prediction.
        The numbers bellow are the counting results. 
    }
    \label{fig:res_fsc}
    \vspace{-13pt}
\end{figure*}

\vspace{2pt}\noindent \textbf{FSC-147.}  Quantitative results on FSC-147 are given in \cref{tab:res_fsc_147}. Our method is compared with GMN~\cite{gmn}, MAML~\cite{maml}, FamNet~\cite{famnet}, and CFOCNet~\cite{cfocnet}. Our approach outperforms all counterparts with a quite large margin. For example, we surpass FamNet+ by $8.47$ MAE and $21.87$ RMSE on validation set, $7.76$ MAE and $14.00$ RMSE on test set. Note that FamNet+ needs test-time adaptation for novel classes, while \textit{our \method needs no test-time adaptation}. These significant advantages demonstrate the effectiveness of our method. In \cref{fig:res_fsc}, we show some qualitative results of \method. Compared with Famnet+~\cite{famnet}, our \method has much stronger ability to separate each independent object within densely packed objects, thus helps obtain an accurate count. Especially, for densely packed green peas (\cref{fig:res_fsc}b), we not only exactly predict the object count, but also localize target objects with such a high precision that every single object could be clearly distinguished.

\begin{table}[tb]
\caption{
    \textbf{Quantitative results} on FSC-147 dataset~\cite{famnet}, where we surpass other competitors by a sufficiently large margin. 
}
\setlength\tabcolsep{8pt}
\centering
\scriptsize
\begin{threeparttable}
\begin{tabular}{lcccc}
\toprule
\multirow{2}{*}{Method}
& \multicolumn{2}{c}{Val Set} & \multicolumn{2}{c}{Test Set} \\
\cmidrule{2-5}
& MAE & RMSE & MAE & RMSE \\
\midrule
GMN~\cite{gmn}         & 29.66 & 89.81 & 26.52 & 124.57 \\
MAML~\cite{maml}       & 25.54 & 79.44 & 24.90 & 112.68 \\
FamNet~\cite{famnet}   & 24.32 & 70.94 & 22.56 & 101.54  \\
FamNet+~\cite{famnet}  & 23.75 & 69.07 & 22.08 & 99.54  \\
CFOCNet~\cite{cfocnet} & 21.19 & 61.41 & 22.10 & 112.71 \\
\midrule
\method (ours) & \textbf{15.28} & \textbf{47.20} & \textbf{14.32} & \textbf{85.54} \\
\bottomrule
\end{tabular}
\end{threeparttable}
\label{tab:res_fsc_147}
\vspace{-10pt}
\end{table}

\vspace{2pt}\noindent \textbf{Cross-validation of FSC-147.}  
The dataset split and shot number are both fixed in FSC-147 benchmark, which could not provide a comprehensive evaluation. 
Therefore, we incorporate FSC-147 with cross-validation to evaluate FSC methods with 4 dataset splits and 2 shot numbers. 
Our approach is compared with FSC baselines including GMN~\cite{gmn} and FamNet~\cite{famnet}. 
These baselines are trained and evaluated by ourselves with the official code. 
The cross-validation results are shown in \cref{tab:fsc_cv}, where fold-$i$ ($i=0,1,2,3$) indicates the test set. 
Under all dataset splits and shot numbers, our method significantly outperforms baseline methods with both MAE and RMSE. 
Averagely, we outperform FamNet by 8.82 MAE and 20.18 RMSE in 1-shot case, 9.67 MAE and 21.74 RMSE in 3-shot case. 
Moreover, from 1-shot case to 3-shot case, our approach gains more performance improvement than two baseline methods, reflecting the superior ability of our \method to utilize multiple support images.

\begin{table*}[!ht]
\caption{
    \textbf{Counting performance with cross-validation setting} on FSC-147 dataset~\cite{famnet}.
    Fold-$i$ ($i=0,1,2,3$) indicates the test set.
    $\rm \Delta$ stands for the averaged improvement of the 3-shot case over the 1-shot case. 
}
\centering
\scriptsize
\begin{threeparttable}
\begin{tabular}{cl|ccccc|ccccc|c}
\toprule
\multirow{2}{*}{Metric} & \multirow{2}{*}{Method} 
& \multicolumn{5}{c|}{1-shot} & \multicolumn{5}{c|}{3-shot} & \multirow{2}{*}{$\rm \Delta$} \\ 
& & Fold-0 & Fold-1 & Fold-2 & Fold-3 & Mean & Fold-0 & Fold-1 & Fold-2 & Fold-3 & Mean & \\ 
\midrule
\multirow{3}{*}{MAE} 
& GMN~\cite{gmn} & 37.44 & 21.89 & 31.52 & 32.73 & 30.90 & 36.53 & 21.43 & 31.23 & 31.51 & 30.18 & -0.72 \\
& FamNet~\cite{famnet} & 27.98 & 15.75 & 21.32 & 22.33 & 21.85 & 26.32 & 15.51 & 21.28 & 21.96 & 21.27 & -0.58 \\
\cmidrule{2-13}
& \method (ours) & \textbf{17.64} & \textbf{6.97} & \textbf{12.96} & \textbf{14.55} & \textbf{13.03} & \textbf{13.21} & \textbf{6.58} & \textbf{12.43} & \textbf{14.16} & \textbf{11.60} & \textbf{-1.43} \\
\midrule
\multirow{3}{*}{RMSE} 
& GMN~\cite{gmn} & 111.68 & 45.75 & 127.94 & 75.45 & 90.21 & 109.31 & 44.44 & 128.77 & 73.76 & 89.07 & -1.14 \\
& FamNet~\cite{famnet} & 86.04 & 34.61 & 101.68 & 53.47 & 68.95 & 76.03 & 33.41 & 107.45 & 50.25 & 66.79 & -2.16 \\
\cmidrule{2-13}
& \method (ours) & \textbf{53.99} & \textbf{16.13} & \textbf{85.28} & \textbf{39.66} & \textbf{48.77} & \textbf{38.94} & \textbf{14.25} & \textbf{88.72} & \textbf{38.30} & \textbf{45.05} & \textbf{-3.72} \\ 
\bottomrule
\end{tabular}
\end{threeparttable}
\vspace{-10pt}
\label{tab:fsc_cv}
\end{table*}

\subsection{Cross-dataset Generalization}

Following FamNet~\cite{famnet}, we test our model’s generalization ability on the car counting dataset, CARPK~\cite{lpn}. The models are first pre-trained on FSC-147~\cite{famnet} (the ``car'' category is excluded), then fine-tuned on CARPK. The results are shown in \cref{tab:res_car_crowd}. For the models pre-trained on FSC-147, we significantly surpass FamNet by 42.23\% in MAE and 45.85\% in RMSE. When it comes to the fine-tuning scenario, our method still consistently outperforms all baselines. For instance, we surpass GMN~\cite{gmn} by 28.74\% in MAE and 28.89\% in RMSE. Therefore, our \method has much better ability in cross-dataset generalization.

\begin{table}[tb]
\caption{
    \textbf{Cross-dataset generalization} on the car counting dataset CARPK~\cite{lpn}. The models are first pre-trained on FSC-147~\cite{famnet} (the ``car'' category is excluded), then fine-tuned on CARPK. 
}
\setlength\tabcolsep{8pt}
\centering
\scriptsize
\begin{threeparttable}
\begin{tabular}{clcc}
\toprule
& Method & MAE & RMSE \\
\midrule
\multirow{3}{*}{\tabincell{c}{Pre-trained on \\ FSC-147~\cite{famnet}}}
& GMN~\cite{gmn}       & 32.92 & 39.88  \\
& FamNet~\cite{famnet} & 28.84 & 44.47  \\
\cmidrule{2-4}
& \method (ours)       & 16.66 & 24.08  \\
\midrule
\multirow{3}{*}{\tabincell{c}{Fine-tuned on \\ CARPK~\cite{lpn}}}
& GMN~\cite{gmn}       & 7.48  & 9.90   \\
& FamNet~\cite{famnet} & 18.19 & 33.66  \\
    \cmidrule{2-4}
& \method (ours)       & \textbf{5.33} & \textbf{7.04}  \\
\bottomrule
\end{tabular}
\end{threeparttable}
\label{tab:res_car_crowd}
\vspace{-15pt}
\end{table}

\subsection{Ablation Study} \label{subsec:ablation}
To verify the effectiveness of the proposed modules and the selection of hyper-parameters, we implement extensive ablation studies on FSC-147~\cite{famnet}. 

\vspace{2pt}\noindent \textbf{SCM and FEM.} 
To demonstrate the effectiveness of the proposed SCM and FEM modules, we conduct diagnostic experiments. 
The substitute of FEM is to concatenate the similarity map and query feature together, then recover the feature dimension through a $1\times1$ convolutional layer. 
To invalidate SCM, we replace the score normalization to a naive maximum normalization (dividing the maximum value). 
The results are shown in \cref{tab:ablation}a. 
Both SCM and FEM are necessary for our \method. 
Specifically, when we replace FEM, the performance drops remarkably by 9.05 MAE in test set. 
This reflects that FEM is of vital significance in our \method, since the core of our insight, \textit{i.e.} feature enhancement, is completed in FEM. 
Besides, when we remove SCM, the performance also drops by 1.81 MAE in test set. 
This indicates that SCM derives a similarity map with a proper value range, promoting the performance.

\begin{table*}[tb]
\caption{
    \textbf{Ablation studies} on (a) the effect of the similarity comparison module (SCM) and feature enhancement module (FEM), (b) the score normalization in SCM, (c) the stacked number of our \method block, and (d) the place to regress density map, (e) comparison with attention, (f) kernel flipping, (g) training or freezing backbone. 
}
\begin{minipage}[t]{0.25\textwidth}
\setlength\tabcolsep{1pt}
\centering
\scriptsize
(a) SCM \& FEM \\
\begin{tabular}{cccccc}
\toprule
\multirow{2}{*}{SCM} & \multirow{2}{*}{FEM} & \multicolumn{2}{c}{Val Set} & \multicolumn{2}{c}{Test Set} \\
\cmidrule{3-6}
& & MAE & RMSE & MAE & RMSE \\
\midrule
\ding{55} & \ding{55} & 21.35 & 62.13 & 22.10 & 99.89 \\
\ding{51} & \ding{55} & 21.45 & 59.15 & 23.37 & 98.01 \\
\ding{55} & \ding{51} & 17.55 & 58.66 & 16.13 & 96.90 \\
\ding{51} & \ding{51} & \textbf{15.28} & \textbf{47.20} & \textbf{14.32} & \textbf{85.54} \\
\bottomrule
\end{tabular}
\end{minipage}
\hfill
\begin{minipage}[t]{0.25\textwidth}
\setlength\tabcolsep{1pt}
\centering
\scriptsize
(b) Score Normalization \\
\begin{tabular}{cccccc}
\toprule
\multirow{2}{*}{ENorm} & \multirow{2}{*}{SNorm} & \multicolumn{2}{c}{Val Set} & \multicolumn{2}{c}{Test Set} \\
\cmidrule{3-6}
& & MAE & RMSE & MAE & RMSE \\
\midrule
\ding{55} & \ding{55} & 17.55 & 58.66 & 16.13 & 96.90 \\
\ding{51} & \ding{55} & 16.55 & 51.87 & 15.14 & 85.65 \\
\ding{55} & \ding{51} & 16.58 & 51.26 & 16.40 & 93.97 \\
\ding{51} & \ding{51} & \textbf{15.28} & \textbf{47.20} & \textbf{14.32} & \textbf{85.54} \\
\bottomrule
\end{tabular}
\end{minipage}
\hfill
\begin{minipage}[t]{0.23\textwidth}
\setlength\tabcolsep{1pt}
\centering
\scriptsize
\vspace{-12.5pt}
(c) Number of Block \\
\begin{tabular}{ccccc}
\toprule
\multirow{2}{*}{\# Block} 
& \multicolumn{2}{c}{Val Set} & \multicolumn{2}{c}{Test Set} \\
\cmidrule{2-5}
& MAE & RMSE & MAE & RMSE \\
\midrule
1 & 16.23 & 55.34  & 16.46 & 92.62  \\
2 & 16.04 & 54.53  & 15.36 & 87.35  \\
3 & 15.78 & 53.39  & 14.74 & 88.22  \\
4 & \textbf{15.28} & \textbf{47.20} & \textbf{14.32} & \textbf{85.54} \\
5 & 15.67 & 50.73 & 15.54 & 96.10 \\
\bottomrule
\end{tabular}
\end{minipage}
\hfill
\begin{minipage}[t]{0.25\textwidth}
\setlength\tabcolsep{1pt}
\centering
\scriptsize
(d) Similarity Map \textit{v.s.} Enhanced Feature \\
\begin{tabular}{lcccc}
\toprule
\multirow{2}{*}{} 
& \multicolumn{2}{c}{Val Set} & \multicolumn{2}{c}{Test Set} \\
\cmidrule{2-5}
& MAE & RMSE & MAE & RMSE \\
\midrule
Raw Simi. & 24.36 & 74.61 & 23.65 & 108.77 \\
1-block Feat. & 16.23 & 55.34 & 16.46 & 92.62 \\
4-block Simi. & 19.74 & 64.30 & 18.70 & 99.34 \\
4-block Feat. & \textbf{15.28} & \textbf{47.20} & \textbf{14.32} & \textbf{85.54} \\
\bottomrule
\end{tabular}
\end{minipage}
\vfill
\vspace{3pt}
\begin{minipage}[t]{0.33\textwidth}
\setlength\tabcolsep{3pt}
\centering
\scriptsize
(e) Vanilla Attention~\cite{attention_need} \textit{v.s.} \method
\begin{tabular}{ccccc}
\toprule
& \multicolumn{2}{c}{Val Set} & \multicolumn{2}{c}{Test Set} \\
\cmidrule{2-5}
& MAE & RMSE & MAE & RMSE \\
\midrule
Vanilla Attention~\cite{attention_need} & 20.45 & 55.22 & 20.21 & 93.47  \\
\method & \textbf{15.28} & \textbf{47.20} & \textbf{14.32} & \textbf{85.54} \\
\bottomrule
\end{tabular}
\end{minipage}
\hfill
\begin{minipage}[t]{0.33\textwidth}
\setlength\tabcolsep{3pt}
\centering
\scriptsize
(f) Kernel Flipping in FEM
\begin{tabular}{ccccc}
\toprule
\multirow{2}{*}{Kernel Flipping}
& \multicolumn{2}{c}{Val Set} & \multicolumn{2}{c}{Test Set} \\
\cmidrule{2-5}
& MAE & RMSE & MAE & RMSE \\
\midrule
\ding{55} & 16.78 & 57.47 & 15.35 & 93.59 \\
\ding{51} & \textbf{15.28} & \textbf{47.20} & \textbf{14.32} & \textbf{85.54} \\
\bottomrule
\end{tabular}
\end{minipage}
\hfill
\begin{minipage}[t]{0.33\textwidth}
\setlength\tabcolsep{3pt}
\centering
\scriptsize
(g) Training \textit{v.s.} Freezing Backbone
\begin{tabular}{ccccc}
\toprule
\multirow{2}{*}{Freezing Backbone}
& \multicolumn{2}{c}{Val Set} & \multicolumn{2}{c}{Test Set} \\
\cmidrule{2-5}
& MAE & RMSE & MAE & RMSE \\
\midrule
\ding{55} & 25.24 & 65.23 & 26.00 & 103.83 \\
\ding{51} & \textbf{15.28} & \textbf{47.20} & \textbf{14.32} & \textbf{85.54} \\
\bottomrule
\end{tabular}
\end{minipage}
\vspace{-8pt}
\label{tab:ablation}
\end{table*}

\begin{figure*}[tb]
    \centering
    \includegraphics[width=0.93\linewidth]{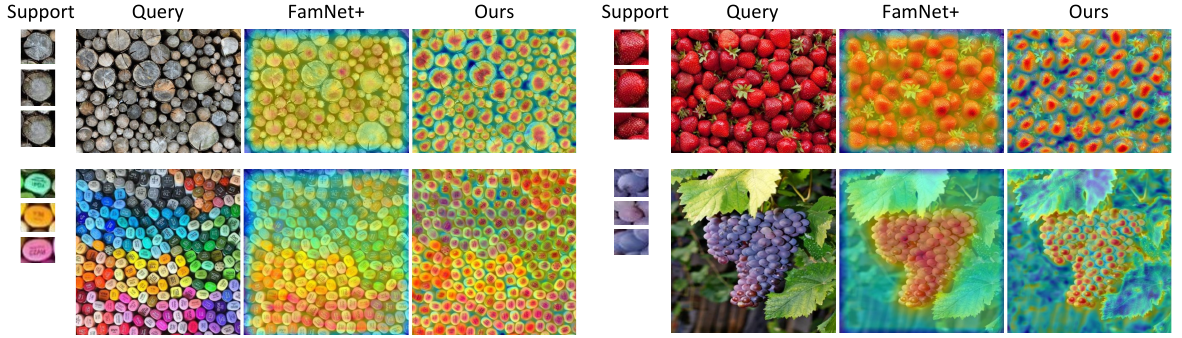}
    \vspace{-2pt}
    \caption{
        \textbf{Visualization of the similarity maps} developed by FamNet+~\cite{famnet} and our \method.
        Benefiting from the proposed \method block, our approach recognizes \textit{much clearer boundaries} between densely packed objects. 
    }
    \label{fig:relation}
    \vspace{-15pt}
\end{figure*}

\vspace{2pt}\noindent \textbf{Score Normalization.} 
We conduct ablation experiments regarding ENorm and SNorm in \cref{tab:ablation}b. 
A naive maximum normalization (dividing the maximum value) serves as the baseline when both normalization methods are removed. 
Even if without score normalization, we still stably outperform all baselines in \cref{tab:res_fsc_147}.  
Adding either ENorm or SNorm improves the performance greatly ($\geq 4$ MAE in test set), indicating the significance of score normalization. 
ENorm together with SNorm brings the best performance, reflecting that the two normalization methods could cooperate together for further performance improvement.

\vspace{2pt}\noindent \textbf{Block Number.} 
It is described in \cref{subsec:framework} that our \method could be formulated to a multi-block architecture. 
Here we explore the influence of the block number. 
As shown in \cref{tab:ablation}c, only 1-block \method has achieved state-of-the-art performance by a large margin, which illustrates the effectiveness of our designed \method architecture. 
Furthermore, the performance gets improved gradually when the block number increases from 1 to 4, and decreased slightly when the block number is added to 5. 
As proven in \cite{resnet}, too many blocks could hinder the training process, decreasing the performance. Finally, we set the block number as 4 for FSC-147.

\vspace{-2pt} \noindent \textbf{Regressing from Similarity Map \textit{v.s.} Enhanced Feature.}  
The density map can be regressed from either the enhanced feature or the similarity map. 
We compare these two choices in \cref{tab:ablation}d. 
\textit{Raw Similarity} is similar to FamNet~\cite{famnet} (without test-time adaptation), predicting the density map directly from the raw similarity. 
The rest 3 methods follow our design, where \textit{$i$-block Similarity} and \textit{$i$-block Feature} mean that the density map is regressed from the similarity map and enhanced feature of the $i^{th}$ block, respectively. 
Obviously, \textit{1-block Feature} and \textit{4-block Feature} significantly outperform \textit{Raw Similarity} and \textit{4-block Similarity}, respectively. 
The reason may be that the enhanced feature contains rich semantics and can filter out some erroneous high similarity values, \textit{i.e.} the high similarity values that do not correspond to target objects, as proven in~\cite{pwcnet}.

\noindent \textbf{Comparison with Attention.} In \cref{tab:ablation}e, when the similarity derivation in SCM and the feature aggregation in FEM are replaced by an vanilla attention~\cite{attention_need}, the performance drops dramatically. As stated in \cref{subsec:core-block}, our method could better utilize the \textit{spatial structure} of features than vanilla attention, which helps find more accurate boundaries between objects and brings substantial improvement.

\noindent\textbf{Kernel Flipping in FEM.}
The kernel flipping in FEM could help the similarity-weighted feature, $\fweight$, inherit the spatial structure from the support feature, $\fs$ (see \supp for details). 
The effectiveness of adding the flipping is proven by \cref{tab:ablation}f. 
Adding the flipping could improve the performance stably ($\geq$ 1 MAE), reflecting that preserving the spatial structure of $\fweight$ benefits the counting performance.

\vspace{2pt}\noindent\textbf{Training \textit{v.s.} Freezing Backbone.} The comparison results are provided in \cref{tab:ablation}g. The frozen backbone significantly surpasses the trainable backbone. Considering that the testing classes are different from training classes in FSC-147~\cite{famnet}, training backbone will lead the backbone to extract more relevant features to training classes, which decreases the performance in the validation and test sets.



\subsection{Visualization}

We visualize and compare the intermediate similarity map in FamNet~\cite{famnet} and \method in \cref{fig:relation}, which intuitively explains why \method surpasses FamNet substantially. 
Here the similarity map in \method means the one in the last block. 
In FamNet, the similarity map is derived by direct comparison between the raw features of the query image and support images. 
However, the similarity map is far less informative than features, making it hard to identify clear boundaries within densely packed objects. 
In contrast, we weigh the support feature based on the similarity values, then integrate the similarity-weighted feature into the query feature. 
This design encodes the support-query relationship into features, while keeping the rich semantics extracted from the image. 
Also, our similarity comparison module is learnable. 
Benefiting from these, our \method gets clear boundaries between densely packed objects in the similarity map, which is beneficial to regress an accurate count.

\section{Conclusion}\label{sec:conclusion}

In this work, to tackle the challenging few-shot object counting task, we propose the similarity-aware feature enhancement block, composed of a similarity comparison module (SCM) and a feature enhancement module (FEM). Our SCM compares the support feature and the query feature to derive a score map. Then the score map is normalized across both the exemplar and spatial dimensions, producing a reliable similarity map. The FEM views these similarity values as weighting coefficients to integrate the support features into the query feature. By doing so, the model will pay more attention to the regions similar to support images, bringing distinguishable borders within densely packed objects. Extensive experiments on various benchmarks and training settings demonstrate that we achieve state-of-the-art performance by a considerably large margin.


{
\small
\clearpage
\bibliographystyle{ieee_fullname}
\bibliography{ref}
}

\appendix
\section*{Appendix}
\section{Network Architecture and Training Configurations}\label{appendix:sec:architecture}

This part describes the detailed architecture of our \method block and other assistant modules, followed by the training configurations.

\vspace{2pt}\noindent\textbf{Feature Extractor.} 
We select ResNet-18~\cite{resnet} pre-trained on ImageNet~\cite{imagenet} as the feature extractor.%
\footnote{We borrow the checkpoint \href{https://download.pytorch.org/models/resnet18-5c106cde.pth}{here}.}
Given a query image, $Q \in \mathbb{R}^{3 \times 512 \times 512}$, we resize the outputs of the first three residual stages of ResNet-18 to the same size, $128 \times 128$, and concatenate them along the channel dimension. 
Afterward, a $1 \times 1$ convolutional layer is applied to reduce the channel dimension to 256, resulting in the query feature, $\fq \in \mathbb{R}^{256 \times 128 \times 128}$. 
The size of ROI pooling~\cite{faster_rcnn} is set as $3 \times 3$, so the support feature, $\fs$, has the shape of $K \times 256 \times 3 \times 3$ in the $K$-shot case. 
The backbone is frozen during training, while the $1 \times 1$ convolutional layer is not.

\vspace{2pt}\noindent\textbf{Similarity Comparison Module (SCM).}
Our SCM is implemented with three steps: \textit{learnable feature projection}, \textit{feature comparison}, and \textit{score normalization}. 
The \textit{feature comparison} is implemented by convoluting the query feature, $\fq$, with the support feature, $\fs$, as kernels, deriving a score map, $\Rraw$. This process is illustrated intuitively in \cref{appendix:fig:conv}a. 
Other components in SCM have been detailed in the paper. 
The SCM finally outputs a similarity map, $\Rnorm \in \mathbb{R}^{K \times 1 \times 128 \times 128}$.

\begin{figure*}[tb]
  \centering
  \includegraphics[width=1.0\linewidth]{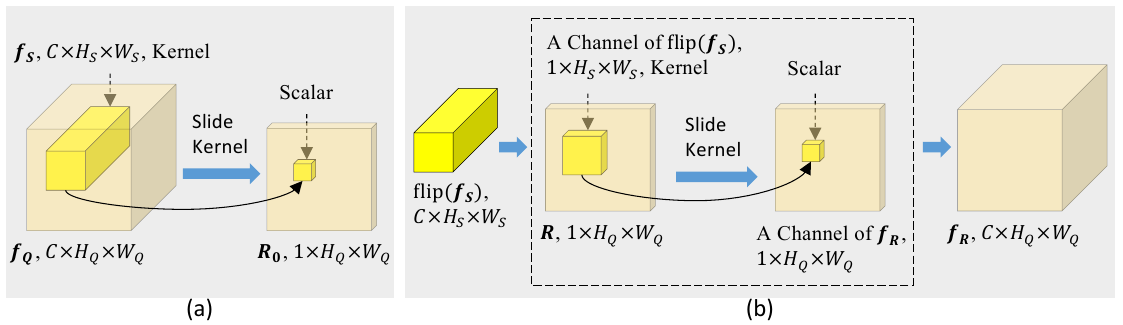}
  \caption{\textbf{(a) Illustration of the \textit{feature comparison}} in SCM under the 1-shot case, where the feature projection is omitted. \textbf{(b) Illustration of the \textit{weighted feature aggregation}} in FEM under the 1-shot case.}
  \label{appendix:fig:conv}
\end{figure*}

\begin{figure}[tb]
\centering
\includegraphics[width=1\linewidth]{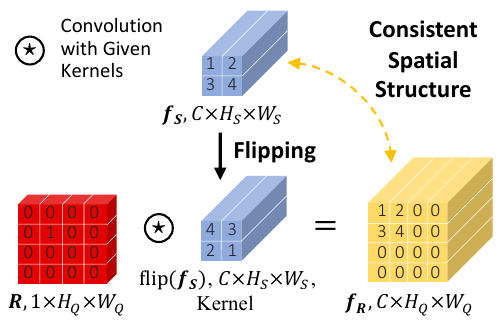}
\caption{
    \textbf{Illustration of kernel flipping in FEM}, which helps $\fweight$ inherit the spatial structure from $\fs$. The convolution is implemented with the same padding strategy. 
}
\label{appendix:fig:flip}
\end{figure}

\vspace{2pt}\noindent\textbf{Feature Enhancement Module (FEM).} 
The FEM is composed of two steps: \textit{weighted feature aggregation} and \textit{learnable feature fusion}. 
The \textit{weighted feature aggregation} treats the values in the similarity map, $\Rnorm$, as weighting coefficients to integrate $\fs$, producing the similarity-weighted feature, $\fweight$. 
This process is realized by the convolution, as shown in \cref{appendix:fig:conv}b. 
Besides, before serving as the convolutional kernels, $\fs$ is flipped both horizontally and vertically. 
As illustrated in \cref{appendix:fig:flip}, the flipping helps $\fweight$ inherit the spatial structure from $\fs$. 
In \cref{appendix:fig:flip}, $\Rnorm$ is a unit impulse function, meaning that only one position has the maximum similarity with $\fs$, while other positions have no similarity with $\fs$ at all. 
Therefore, $\fweight$ should have a sub-part exactly the same with $\fs$ in the position corresponding to the maximum similarity, while the others should be zero vectors. 
The \textit{weighted feature aggregation} constructs $\fweight$ following the above insights via flipping and convolution. 
The \textit{learnable feature fusion} is completed by a 2-layer convolutional network, skip connection, and layer normalization. 
The architecture of the convolutional network is shown in \cref{appendix:tab:conv}a. 
Other components in FEM have been detailed in the paper. 
Eventually, the FEM produces the enhanced feature, $\fout \in \mathbb{R}^{256 \times 128 \times 128}$.

\begin{table}[t]
    \caption{
        \textbf{Network architectures} of (a) the \textit{learnable feature fusion} in FEM, where the skip connection and the layer normalization are omitted, and (b) the regress head. 
    }
    \vspace{8pt}
    \label{appendix:tab:conv}
    \scriptsize
    \begin{minipage}[t]{0.5\textwidth}
    \setlength\tabcolsep{8pt}
    \centering
    (a) Architecture of the learnable feature fusion \\
    \vspace{3pt}
    \begin{tabular}{ccccc}
        \toprule
        layer & kernel & in & out & activation \\
        \midrule
        Conv   & $3 \times 3$ & 256 & 1024 & Leaky ReLU \\
        \midrule
        Conv   & $3 \times 3$ & 1024 & 256 & - \\
        \bottomrule
    \end{tabular}
    \end{minipage}
    \vfill
    \vspace{8pt}
    \begin{minipage}[t]{0.5\textwidth}
    \setlength\tabcolsep{6pt}
    \centering
    (b) Architecture of the regress head \\
    \vspace{3pt}
    \begin{tabular}{cccccc}
    \toprule
        layer & kernel & in & out & activation & followed by \\
        \midrule
        Conv & $5 \times 5$ & 256 & 128 & Leaky ReLU & $2 \times$Upsample \\
        \midrule
        Conv & $3 \times 3$ & 128 & 64 & Leaky ReLU & $2 \times$Upsample \\
        \midrule
        Conv & $1 \times 1$ & 64 & 32 & Leaky ReLU & - \\
        \midrule
        Conv & $1 \times 1$ & 32 & 1  & ReLU & - \\
        \bottomrule
    \end{tabular}
    \end{minipage}
\end{table}

\vspace{2pt}\noindent\textbf{Regress Head.} The regress head regresses the density map, $\bm{D} \in \mathbb{R}^{512 \times 512}$, from the enhanced feature, $\fout$. 
The regression head is composed of a sequence of convolutional layers, followed by the Leaky ReLU activation and bi-linear upsampling, as shown in \cref{appendix:tab:conv}b.

\vspace{2pt}\noindent\textbf{Multi-block Architecture.} The enhanced feature derived by one block, $\fout$, could serve as the input to the next block by taking the place of the query feature, $\fq$, forming a multi-block architecture. As for another input of the next block, the support feature, $\fs$, there are two choices. If the support image is cropped from the query image, $\fs$ is updated by ROI pooling on newly obtained $\fout$. If not, $\fs$ does not change in different blocks.

\vspace{2pt}\noindent\textbf{Training Configuration on FSC-147 \cite{famnet}.} 
The sizes of the query image, the query feature, and the support feature are selected as $512 \times 512$, $128 \times 128$, and $3 \times 3$, respectively. 
The \method block number is set as 4. 
The model is trained with Adam optimizer \cite{adam} for 200 epochs with batch size 8. 
The hyper-parameter $\epsilon$ in Adam optimizer is set as 4e-11, much smaller than the default 1e-8, considering the small norm of the losses and the gradients. 
The learning rate is set as 2e-5 initially, and it is dropped by 0.25 after every 80 epochs. 
Data augmentation methods including random horizontal flipping, color jittering, and random gamma transformation are adopted.

\section{Ablation Studies}\label{appendix:sec:ablation}

This part conducts comprehensive ablation studies on the components of our approach.

\vspace{2pt}\noindent\textbf{Loss Function.}
The loss function described in the paper is MSE loss. Actually, we also implement experiments with another SSIM term as follows,
\begin{align}
    \mathcal{L} = \mathtt{MSE}(\bm{D}, \bm{D_{GT}}) - \alpha \mathtt{SSIM}(\bm{D}, \bm{D_{GT}}), \label{equ:loss}
\end{align}
where $\mathtt{SSIM}(\cdot)$ is the structural similarity function \cite{ssim}, which measures the local pattern consistence between the predicted density map and the ground-truth, $\alpha$ is the weight term. 
The results with different $\alpha$ are shown in \cref{appendix:tab:ablation}a. 
Adding the SSIM term promotes the performance of MAE but with the sacrifice of RMSE. 
Compared with MAE, RMSE relies more heavily on the prediction of the samples with extremely large count. 
Therefore, we speculate that the SSIM term is beneficial to some samples, but may harm the samples with extremely large count. 
We finally decide not to add the SSIM term, because the performance drop of RMSE is too large.

\begin{table}[t]
    \caption{\textbf{Ablation studies} regarding (a) loss weight term $\alpha$ in \cref{equ:loss}, (b) size of ROI pooling.}
    \vspace{5pt}
    \label{appendix:tab:ablation}
    \centering
    \scriptsize
    (a) $\alpha$ in \cref{equ:loss} \\
    \begin{minipage}[t]{0.5\textwidth}
        \centering
        \setlength\tabcolsep{8pt}
        \begin{tabular}{ccccc}
            \toprule
            \multirow{2}{*}{$\alpha$} 
            & \multicolumn{2}{c}{Val Set} & \multicolumn{2}{c}{Test Set} \\
            \cmidrule{2-5}
            & MAE & RMSE & MAE & RMSE \\
            \midrule
            1e-3 & 15.15 & 52.02 & 15.42 & 95.77 \\
            1e-4 & \textbf{14.18} & 53.65 & \textbf{13.55} & 89.69 \\
            1e-5 & 15.11 & 56.05 & 14.63 & 93.41 \\
            0    & 15.28 & \textbf{47.20} & 14.32 & \textbf{85.54} \\
            \bottomrule
        \end{tabular}
    \end{minipage}
    \vfill
    \vspace{5pt}
    (b) Size of ROI Pooling \\
    \begin{minipage}[t]{0.5\textwidth}
        \centering
        \setlength\tabcolsep{6pt}
        \begin{tabular}{ccccc}
            \toprule
            \multirow{2}{*}{Size of ROI Pooling} 
            & \multicolumn{2}{c}{Val Set} & \multicolumn{2}{c}{Test Set} \\
            \cmidrule{2-5}
            & MAE & RMSE & MAE & RMSE \\
            \midrule
            $1 \times 1$ & 15.83 & 54.65  & 16.13 & 95.52 \\
            $3 \times 3$ & \textbf{15.28} & \textbf{47.20} & \textbf{14.32} & \textbf{85.54} \\
            $5 \times 5$ & 15.57 & 53.79 & 15.18 & 89.32 \\
            \bottomrule
        \end{tabular}
    \end{minipage}
\end{table}

\vspace{2pt}\noindent\textbf{Size of ROI Pooling.} 
To study the influence of the ROI pooling size, we conduct experiments with different ROI pooling sizes. 
The results are shown in \cref{appendix:tab:ablation}b. 
The performance is the worst with the ROI pooling size as $1 \times 1$, \textit{i.e.} pooling to a support vector, since pooling to a support vector fully omits the spatial information of the support image.
Adding the ROI pooling size to $3 \times 3$ brings stable improvement. 
However, further increasing the ROI pooling size to $5 \times 5$ decreases the performance slightly. 
This may be because too large ROI pooling size would slightly hinder the accurate localization of target objects. 
Accordingly, we select the ROI pooling size as $3 \times 3$ for FSC-147.

\section{More Results}\label{appendix:sec:results}



This part presents more experimental results, including the quantitative evaluation on various class-specific counting datasets~\cite{lpn, ucsd, mall, mcnn}, as well as some visual samples.

\subsection{Class-specific Object Counting}

Our method is designed to be a general class-agnostic FSC approach. Nonetheless, we still evaluate our method on class-specific counting tasks to further testify its superiority. 

\vspace{2pt}\noindent \textbf{Class-specific Counting Datasets.} We select five class-specific counting datasets including two car counting datasets: CARPK~\cite{lpn} and PUCPR+~\cite{lpn} and three crowd counting datasets: ShanghaiTech (PartA and PartB)~\cite{mcnn}, UCSD~\cite{ucsd}, and Mall~\cite{mall}. The details of these datasets are given in \cref{appendix:tab:datasets}.

\begin{table}[h]
\begin{minipage}[t]{0.5\textwidth}
\setlength\tabcolsep{6pt}
\centering
\scriptsize
\caption{Class-specific counting datasets.}
\vspace{5pt}
\begin{threeparttable}
\begin{tabular}{cccc}
\toprule
Type & Dataset & \#Images & \#Objects \\
\midrule
\multirow{2}{*}{Car}
& CARPK~\cite{lpn} & 1448 & 89,777 \\
& PUCPR+~\cite{lpn} & 125 & 16,916 \\
\midrule
\multirow{4}{*}{Crowd}
& PartA~\cite{mcnn} & 482 & 241,677 \\
& PartB~\cite{mcnn} & 716 & 88,488 \\
& UCSD~\cite{ucsd} & 2000 & 49,885 \\
& Mall~\cite{mall} & 2000 & 62,325 \\
\bottomrule
\end{tabular}
\end{threeparttable}
\label{appendix:tab:datasets}
\end{minipage}
\end{table}

\vspace{2pt}\noindent\textbf{Training Configuration on Class-specific Counting.}
The size of the support feature is set as $1 \times 1$. 
The block number is set as 2. 
Data augmentation methods including random flip, color jitter, random rotation, and random grayscale are used to prevent over-fitting and improve the generalization ability. 
Other setups are the same as \textbf{FSC-147}.

\begin{table}[t]
\caption{
    \textbf{Counting performance on class-specific datasets}, including CARPK~\cite{lpn}, PUCPR+~\cite{lpn}, UCSD~\cite{ucsd}, Mall~\cite{mall}, and ShanghaiTech (Part A \& Part B)~\cite{mcnn}. 
}
\vspace{5pt}
\begin{minipage}[t]{0.5\textwidth}
\centering
\scriptsize
\setlength\tabcolsep{8pt}
(a) Car Counting \\
\begin{threeparttable}
\begin{tabular}{clcccc}
\toprule
\multirow{2}{*}{} & \multirow{2}{*}{Method} & 
\multicolumn{2}{c}{CARPK} & \multicolumn{2}{c}{PUCPR+} \\
\cmidrule{3-6}
& & MAE & RMSE & MAE & RMSE \\
\midrule
\multirow{4}{*}{1} 
& YOLO~\cite{yolo}                 & 48.89 & 57.55 & 156.00 & 200.42 \\
& F-RCNN~\cite{faster_rcnn}        & 47.45 & 57.39 & 111.40 & 149.35 \\
& S-RPN~\cite{lpn}                 & 24.32 & 37.62 & 39.88 & 47.67 \\
& RetinaNet~\cite{retinanet}       & 16.62 & 22.30 & 24.58 & 33.12 \\
   \cmidrule{1-6}
\multirow{2}{*}{2}
& LPN~\cite{lpn}                   & 23.80 & 36.79 & 22.76 & 34.46 \\
& HLCNN~\cite{hlcnn}              & 2.12  & 3.02  & 2.52  & 3.40 \\
    \cmidrule{1-6}
\multirow{3}{*}{4}
& One Look~\cite{one_look}        & 59.46 & 66.84 & 21.88 & 36.73 \\
& IEP Count~\cite{iep_counting}   & 51.83 &   -   & 15.17 &   -   \\
& PDEM~\cite{pdem}                & 6.77  & 8.52  & 7.16  & 12.00 \\
   \cmidrule{1-6}
\multirow{3}{*}{5} 
& GMN~\cite{gmn}                   & 7.48  & 9.90  &   -   &   -   \\
& FamNet~\cite{famnet}             & 18.19 & 33.66 & 14.68\tnote{\dag} & 19.38\tnote{\dag} \\
    \cmidrule{2-6}
& Ours & \textbf{5.33}  & \textbf{7.04} & \textbf{2.42} & \textbf{3.55} \\
\bottomrule
\end{tabular}
\end{threeparttable}
\end{minipage}
\vfill
\vspace{5pt}
\begin{minipage}[t]{0.5\textwidth}
\centering
\scriptsize
\setlength\tabcolsep{8pt}
(b) Crowd Counting (MAE) \\
\begin{threeparttable}
\begin{tabular}{clcccc}
\toprule
 & Method & UCSD & Mall & PartA & PartB \\
\midrule
\multirow{8}{*}{3}
& Crowd CNN~\cite{crowd_cnn}   & 1.60    & -        & 181.8     & 32.0     \\
& MCNN~\cite{mcnn}             & \textbf{1.07} & -  & 110.2     & 26.4     \\
& Switch-CNN~\cite{switch_cnn} & 1.62    &          & 90.4      & 21.6     \\
& CP-CNN~\cite{cp_cnn}         & -       & -        & 73.6      & 20.1     \\
& CRSNet~\cite{crsnet}         & 1.16    & -        & 68.2      & 10.6     \\
& RPNet~\cite{rpnet}           & -       & -        & \textbf{61.2} & 8.1      \\
& GLF~\cite{glf}               & -       & -        & 61.3          & \textbf{7.3}      \\
\cmidrule{1-6}
\multirow{2}{*}{4}
& LC-FCN8~\cite{blobs}         & 1.51      & 2.42      & -           & 13.14    \\
& LC-PSPNet~\cite{blobs}       & 1.01      & 2.00      & -           & 21.61    \\
\cmidrule{1-6}
\multirow{3}{*}{5}
& GMN~\cite{gmn}               & -         & -         & 95.8        & -        \\
& FamNet~\cite{famnet}         & 2.70\tnote{\dag} & 2.64\tnote{\dag} & 159.11\tnote{\dag} & 24.90\tnote{\dag} \\
\cmidrule{2-6}
& Ours & \textbf{0.98} & \textbf{1.69} & \textbf{73.70} & \textbf{9.98}     \\
\bottomrule
\end{tabular}
\end{threeparttable}
\end{minipage}
\begin{threeparttable}
\begin{tablenotes}
\scriptsize
\item[1] Detectors provided by the benchmark~\cite{lpn}.
\item[2] Single-class car counting methods. 
\item[3] Single-class crowd counting methods. 
\item[4] Multi-class counting methods (classes for training and test must be the same). 
\item[5] Few-shot counting methods. 
\item[$\dag$] trained and evaluated by ourselves with the official code.
\end{tablenotes}
\end{threeparttable}
\label{appendix:tab:res_car_crowd}
\end{table}

\vspace{2pt} \noindent \textbf{Car Counting.} Car counting tasks are conducted on CARPK~\cite{lpn} and PUCPR+~\cite{lpn}. 5 support images are randomly sampled from the training set and \textit{fixed for both training and test}. Our method is compared with 4 categories of baselines: object detectors, single-class car counting methods, multi-class counting methods, and FSC methods. Note that multi-class counting methods could only count classes in training set, while FSC methods can count unseen classes. The quantitative results are shown in \cref{appendix:tab:res_car_crowd}a. Our approach surpasses all multi-class counting methods and FSC methods with a large margin, and achieves comparable performance with single-class car counting methods. 

\vspace{2pt} \noindent \textbf{Crowd Counting.} Crowd counting tasks are implemented on UCSD~\cite{ucsd}, Mall~\cite{mall}, and ShanghaiTech~\cite{mcnn}. We randomly sample 5 support images from the training set and \textit{fixed them for both training and test}. 3 kinds of competitors are included: single-class crowd counting methods, multi-class counting methods, and FSC methods. The results of MAE are reported in \cref{appendix:tab:res_car_crowd}b. For UCSD and Mall where the crowd is relatively sparse, our approach surpasses all counterpart methods stably. For ShanghaiTech, our approach outperforms all multi-class counting methods and FSC methods with a large margin, and achieves competitive performance on par with specific crowd counting methods. It is emphasized that, our method is not tailored to the specific crowd counting task, while the compared methods are.

\subsection{More Qualitative Results}

\begin{figure*}[htbp]
\begin{minipage}{1\linewidth}
\centering
\includegraphics[width=0.8\linewidth]{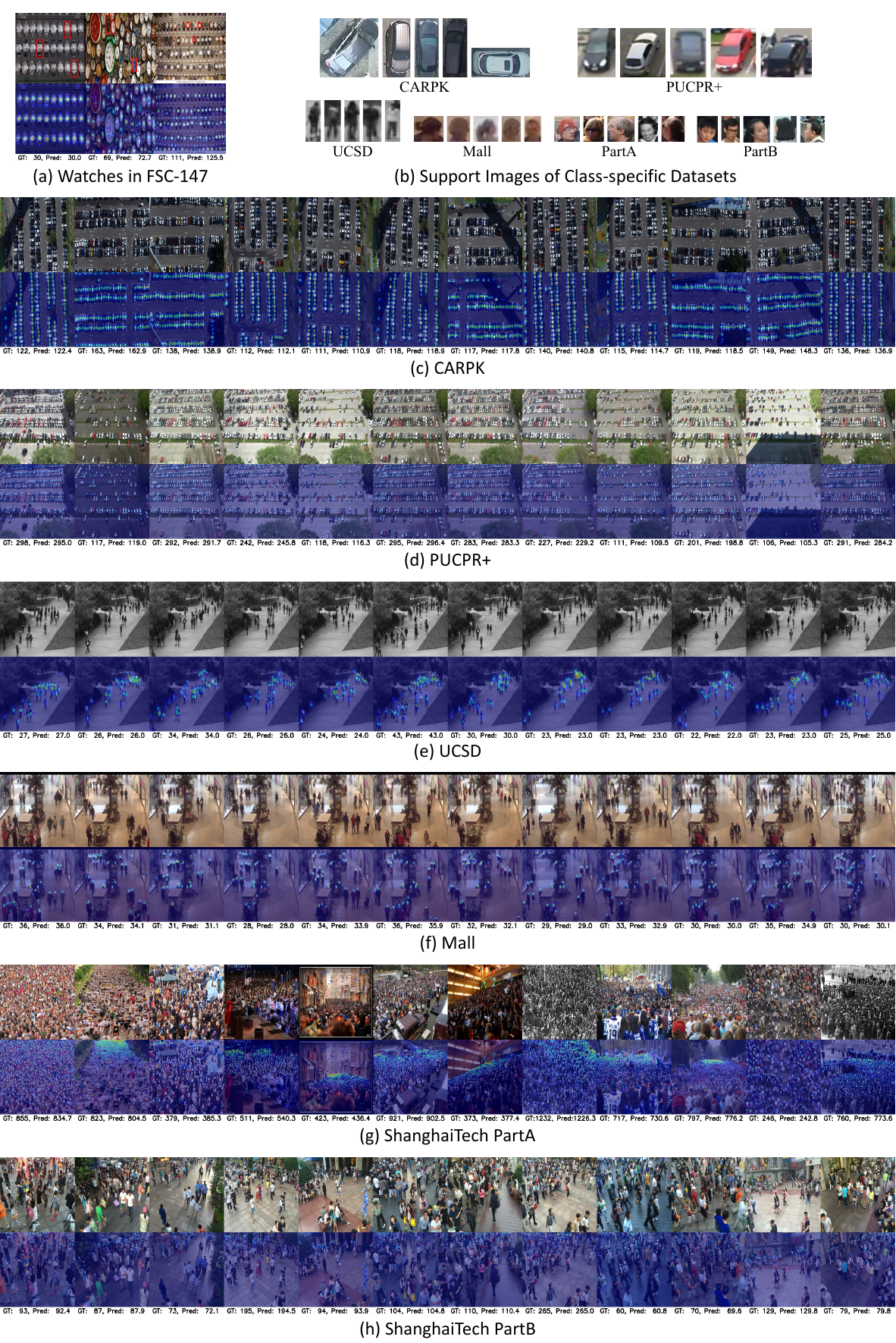}
\end{minipage}
\vspace{-2pt}
\caption{
    (a) Qualitative results on unseen classes (Watches) in FSC-147.
    (b) Support images of class-specific datasets.
    (c-h) Qualitative results on class-specific datasets.
}
\label{appendix:fig:res_1cls}
\end{figure*}

\vspace{2pt}\noindent\textbf{Qualitative Results on FSC-147 \cite{famnet}.} 
The qualitative results of FSC-147 are shown in  \cref{appendix:fig:fsc_1}, \cref{appendix:fig:fsc_2}, and \cref{appendix:fig:res_1cls}a. 
For each class, the images from top to down are the query image and the predicted density map. 
The objects circled by the red rectangles are the support images. 
The texts below the density map describe the counting results. 
Our \method could successfully count objects of all categories with various densities and scales, demonstrating strong generalization ability and robustness. 
Specifically, for both objects with extremely high density (\textit{e.g.}, Legos in Fig. \ref{appendix:fig:fsc_2}) and objects with quite sparse density (\textit{e.g.}, Prawn Crackers in Fig. \ref{appendix:fig:fsc_2}), both small objects (\textit{e.g.}, Birds in Fig. \ref{appendix:fig:fsc_1}) and large objects (\textit{e.g.}, Horses in Fig. \ref{appendix:fig:fsc_1}), both round objects (\textit{e.g.}, Apples in Fig. \ref{appendix:fig:fsc_1}) and square objects (\textit{e.g.}, Stamps in Fig. \ref{appendix:fig:fsc_2}), both vertical strip objects (\textit{e.g.}, Skis in Fig. \ref{appendix:fig:fsc_2}) and horizontal strip objects (\textit{e.g.}, Shirts in Fig. \ref{appendix:fig:fsc_2}), our approach could precisely count objects of interest with high localization accuracy.

\vspace{2pt}\noindent\textbf{Qualitative Results on Class-specific Object Counting.}
Our method is evaluated on two car counting datasets and three crowd counting datasets. 
For each dataset, five support images are randomly sampled from the training set and fixed for both training and test, as shown in \cref{appendix:fig:res_1cls}b. 
The qualitative results on CARPK~\cite{lpn}, PUCPR+~\cite{lpn}, UCSD~\cite{ucsd}, Mall~\cite{mall}, and ShanghaiTech~\cite{mcnn} are shown in \cref{appendix:fig:res_1cls}c-h. 
\textit{(1) Car Counting:} Our approach could localize and count cars with different angles and scales successfully. 
Especially, in the cases that some cars are in the deep shadows (\textit{e.g.}, the $7^{th}, 11^{th}$ examples in \cref{appendix:fig:res_1cls}c, the $11^{st}$ example in \cref{appendix:fig:res_1cls}d) or partly hidden under the trees (\textit{e.g.}, the $3^{rd}, 10^{th}$ examples in \cref{appendix:fig:res_1cls}c, the $5^{th}, 12^{nd}$ examples in \cref{appendix:fig:res_1cls}d), our method still accurately localizes these cars, indicating the superiority of our approach.
\textit{(2) Crowd Counting:} In the cases of UCSD and Mall where the crowd density is relatively sparse, our approach could count the number of persons precisely with extremely small error. 
For ShanghaiTech PartA, if the persons in the crowd are distinguishable (\textit{e.g.}, the $1^{st}, 11^{th}$ examples in \cref{appendix:fig:res_1cls}g), our model could localize each person precisely. If the persons are too crowded to distinguish (\textit{e.g.}, the $2^{nd}, 5^{th}$ examples in \cref{appendix:fig:res_1cls}g), our method could predict an accurate density estimate for crowds. 
For ShanghaiTech PartB where most persons are distinguishable, our approach successfully localizes and counts persons, indicating that our approach is capable of crowd counting with various crowd densities.

\begin{figure*}[htbp]
\begin{minipage}{1\linewidth}
\centering
\includegraphics[width=0.8\linewidth]{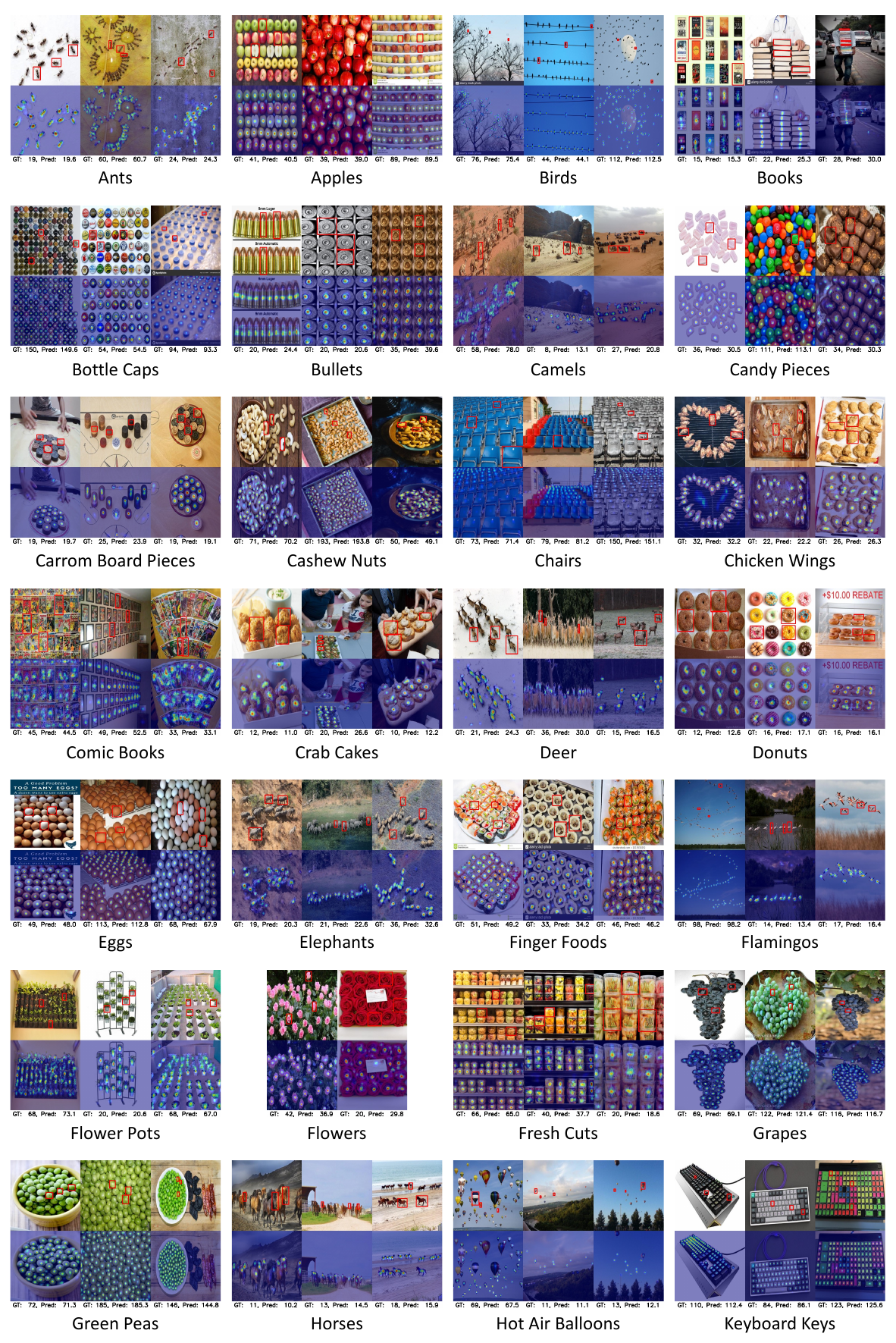}
\end{minipage}
\vspace{-2pt}
\caption{\textbf{Qualitative results} on unseen classes in FSC-147 (from Ants to Keyboard Keys). There are only 2 images of Flowers in FSC-147.}
\label{appendix:fig:fsc_1}
\end{figure*}

\begin{figure*}[htbp]
\begin{minipage}{1\linewidth}
\centering
\includegraphics[width=0.8\linewidth]{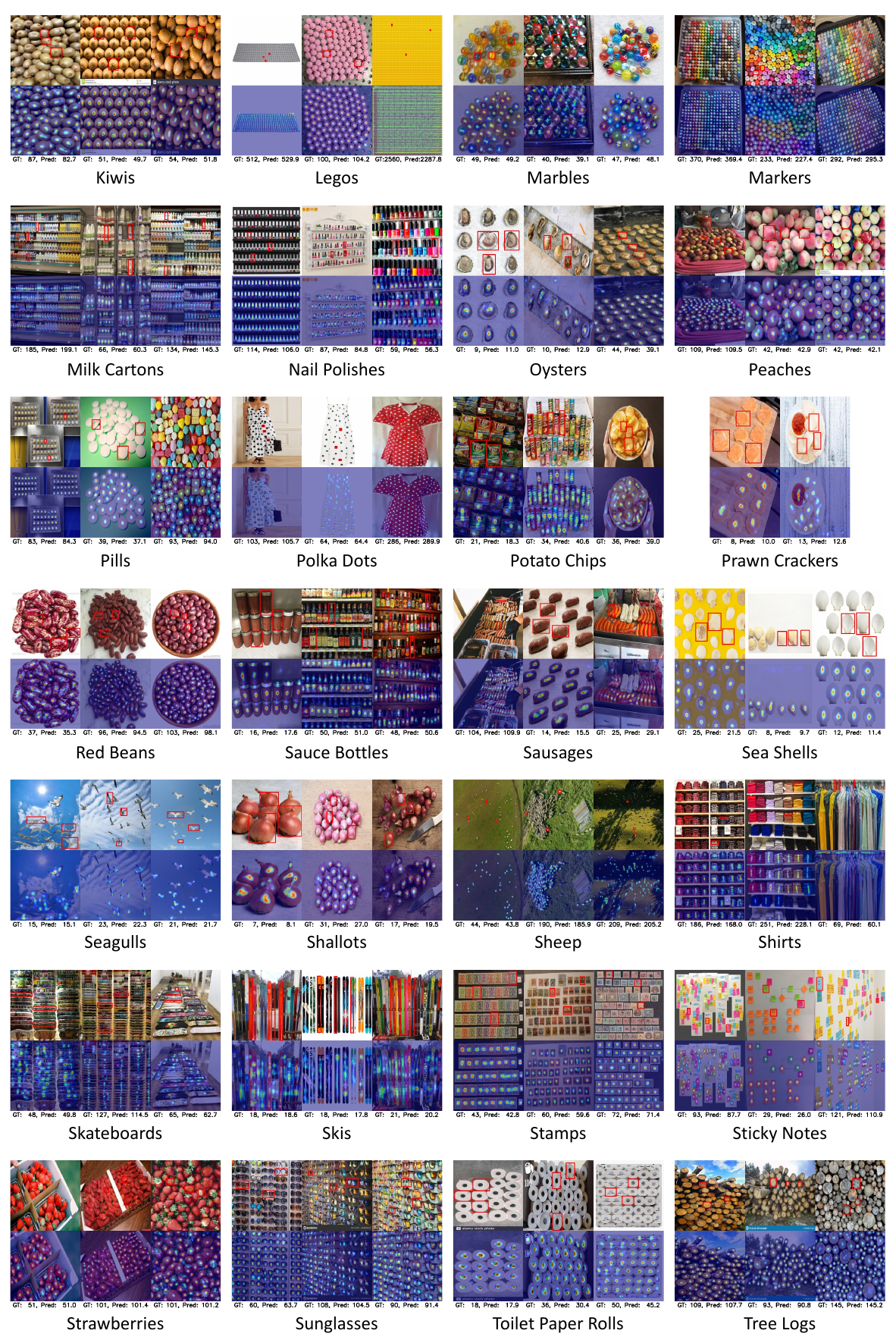}
\end{minipage}
\vspace{-2pt}
\caption{\textbf{Qualitative results} on unseen classes in FSC-147 (from Kiwis to Tree Logs). There are only 2 images of Prawn Crackers in FSC-147.}
\label{appendix:fig:fsc_2}
\end{figure*}

\end{document}